\documentclass{article}
\usepackage{spconf,amsmath,graphicx}
\usepackage{url}

\title{CSPRD: A Financial Policy Retrieval Dataset \\for Chinese Stock Market}
\name{
    Jinyuan Wang\textsuperscript{\rm 1}\sthanks{This work was supported by National Key R\&D Program of China (2021YFC3340700).}, 
    Hai Zhao\textsuperscript{\rm 2}\sthanks{Corresponding author.}, Zhong Wang\textsuperscript{\rm 3}, 
    Zeyang Zhu\textsuperscript{\rm 3}, 
    Jinhao Xie\textsuperscript{\rm 3} \\
    \textit{Yong Yu}\textsuperscript{\rm 3}, 
    \textit{Yongjian Fei}\textsuperscript{\rm 3}, 
    \textit{Yue Huang}\textsuperscript{\rm 3} and \textit{Dawei Cheng}\textsuperscript{\rm 4}
}

\address{ 
	\textsuperscript{\rm 1}SJTU-Paris Elite Institute of Technology, Shanghai Jiao Tong University\\
	\textsuperscript{\rm 2}Department of Computer Science and Engineering, Shanghai Jiao Tong University\\
	\textsuperscript{\rm 3}Shanghai Stock Exchange Technology Co., Ltd.\\
    \textsuperscript{\rm 4}Department of Computer Science and Technology, Tongji University\\
}

\newcommand{\etal}{\textit{et al}. }

\begin{document}
\maketitle
\begin{abstract}
In recent years, great advances in pre-trained language models (PLMs) have sparked considerable research focus and achieved promising performance on the approach of dense passage retrieval, which aims at retrieving relative passages from massive corpus with given questions. However, most of existing datasets mainly benchmark the models with factoid queries of general commonsense, while specialised fields such as finance and economics remain unexplored due to the deficiency of large-scale and high-quality datasets with expert annotations. In this work, we propose a new task, policy retrieval, by introducing the Chinese Stock Policy Retrieval Dataset (CSPRD), which provides 700+ prospectus passages labeled by experienced experts with relevant articles from 10k+ entries in our collected Chinese policy corpus. Experiments on lexical, embedding and fine-tuned bi-encoder models show the effectiveness of our proposed CSPRD yet also suggests ample potential for improvement. Our best performing baseline achieves 56.1\% MRR@10, 28.5\% NDCG@10, 37.5\% Recall@10 and 80.6\% Precision@10 on dev set.
\end{abstract}

\section{Introduction}
\label{sec:intro}


\begin{figure}[t]
    \centering
    \includegraphics[scale=0.35]{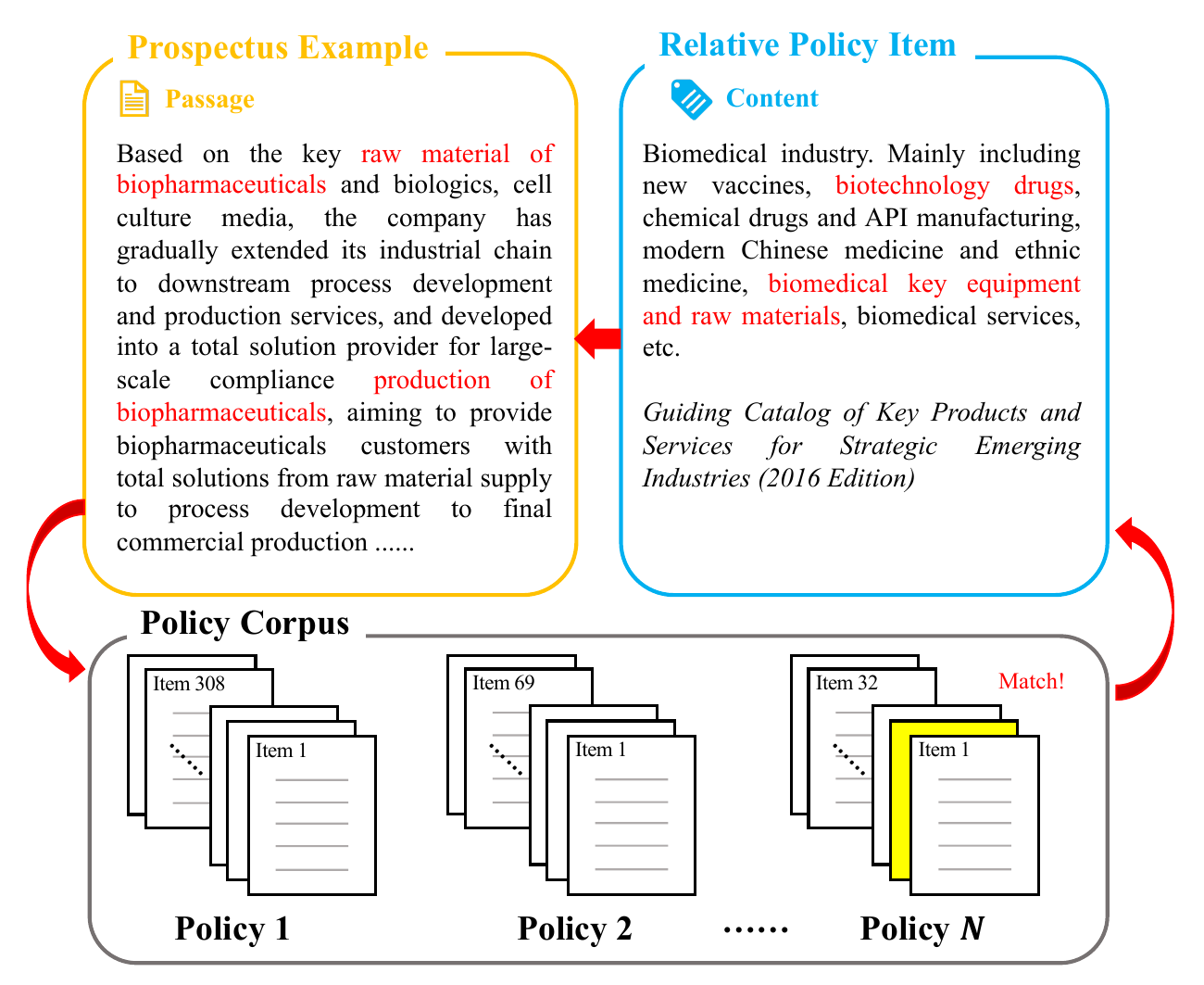}
    \caption{Illustration of the policy retrieval task performed on the Chinese Stock Policy Retrieval Dataset (CSPRD)
    , which consists of 700+ prospectus passages carefully labeled by experienced experts with references to relevant policy articles collected by the Shanghai Stock Exchange.}
    \label{fig:intro}
\end{figure}

Recent advances in pre-trained language models (PLMs) have sparked remarkable success in numerous NLP tasks  \cite{devlin2018bert}. Building upon PLM-based encoders, dense retrieval \cite{karpukhin2020dense, xiong2020approximate} has been proven an effective paradigm for open-domain question-answering \cite{qu2020rocketqa, liu2022retromae}, which aims to retrieve correlated passages of a given query from a massive corpus. However, most existing retrieval datasets substantially benchmark models on general commonsense retrieval \cite{nguyen2016ms}, while specialized domains such as finance and economics remain unexplored due to deficiency of large-scale high-quality datasets with expert annotations \cite{louis2021statutory}. To the best of our knowledge, we are the first work to introduce a policy retrieval dataset, namely the Chinese Stock Policy Retrieval Dataset (CSPRD), filling the blank of fact-driven retrieval dataset in financial and stock market. This domain scenario is extremely sensitive to the reliability of fact, which can also fully test the practical performance of the PLMs.

In this work, we propose a new retrieval task, namely stock policy retrieval, which requires a set of matched policy articles from a large corpus given a passage concerning the primary business in a company’s prospectus. A qualified policy retrieval system can not only provide professional auxiliary services for regulatory agencies, but also provide investors with more thorough information for investment decisions. However, finding policy articles that match the given business description can be a rigorous task as there are two key challenges. (1) Different from general commonsense retrieval \cite{nguyen2016ms}, stock policy retrieval needs to deal with two types of differently distributed languages \cite{louis2021statutory}: complex yet plain language for prospectuses and concise yet fragmentary language for policies. Addressing such difference indirectly demands an almighty system that can not only focus on key information in complex scenario but translate concise expressions into natural language as well. (2) The prospectus passages include vague industry identification and specific product description of the company. However, a policy item match is not solely relied on the accordance of industrial category, but rather on the consistency of business products.

Therefore, a large-scale policy retrieval dataset with expert annotations is necessary to study the extent to which retrieval models can pair with the wisdom and decision-making of professional analysts in regulatory agencies. Admittedly, government policies and prospectuses of listed companies are publicly accessible. However, the review process by regulatory agencies is often a black box, and the matched policy articles of listed companies are not publicly available, which set a high barrier to collecting such datasets. The main contributions of this work are: 

$\bullet$ We introduce the Chinese Stock Policy Retrieval Dataset (CSPRD), which contains a Chinese policy corpus of 10,002 articles and 709 prospectus examples from 545 companies listed on China’s Science and Technology Innovation Board (STAR Market). CSPRD is bilingual in Chinese and English\footnote{Translated by ChatGPT (\texttt{gpt-3.5-turbo-16k-0613})} and is annotated by experienced experts from Shanghai Stock Exchange (SSE). 

$\bullet$ We establish strong baselines on the CSPRD dataset by benchmarking several state-of-the-art retrieval approaches, including lexical, embedding and fine-tuning models. Our best performing baseline achieves 56.1\% MRR@10, 28.5\% NDCG@10, 37.5\% Recall@10 and 80.6\% Precision@10 which shows the effectiveness of our proposed CSPRD yet also suggests ample potential for improvement. 

Our dataset is publicly available on GitHub\footnote{\url{https://github.com/noewangjy/csprd_dataset}}.

\section{Related Work}

\subsection{Specialised Financial Datasets}

Recently, more and more domain-specific datasets are introduced in the NLP community to enrich fact-driven datasets in financial tasks, such as financial sentiment analysis \cite{daudert-etal-2018-leveraging}, numerical reasoning \cite{chen2021finqa} and multilingual topic classification \cite{jorgensen-etal-2023-multifin}.

Some existing works focus on textual information published by enterprises. Daudert \etal \cite{daudert-ahmadi-2019-cofif} introduced CoFiF dataset, which contains 2655 french reports in span of 20 years, covering reference documents, annual, semestrial and trimestrial reports. The JOCo corpus \cite{handschke-etal-2018-corpus} is composed of corporate annual and social responsibility reports of top-ranked international companies. In terms of policy corpus, Wilson \etal  \cite{wilson-etal-2016-creation} created a corpus of 115 privacy policies with manual annotations for fine-grained data practices. However, there is few attention on the policy compliance of enterprise prospectus. To fill this blank, this work is devoted to introducing a Chinese policy retrieval datasets with expert annotations in stock market.


\subsection{Retriever Models}

In general, a policy retriever model is a function that takes a prospectus passage as input along with a corpus of policy articles and returns a small set of relevant policies. Lexical approaches have been the traditional de facto standard for textual retrieval for their robustness and efficiency, such as BM25 \cite{robertson1995okapi} and TF-IDF. In recent years, dense retrieval methods \cite{karpukhin2020dense, xiong2020approximate} have been proven an effective paradigm in open-domain question-answering, which are built upon PLMs-based encoders. Karpukhin \etal \cite{karpukhin2020dense} proposed DPR, which employs a bi-encoder design with in-batch contrastive learning training. Xiong \etal \cite{xiong2020approximate} proposed ANCE, a learning mechanism that selects hard training negatives globally from the entire corpus. Furthermore, retrieval-oriented pre-training methods \cite{qu2020rocketqa, liu2022retromae} are proposed to generate better textual encoding for textual retrieval. Among them, RocketQA \cite{qu2020rocketqa} proposes optimized training approaches to address discrepancy between training and inference. RetroMAE \cite{liu2022retromae} adopt asymmetric masked encoder-decoder design and unbalanced masking ratios during pre-training.



\begin{figure*}[t]
    \centering
    \includegraphics[scale=0.46]{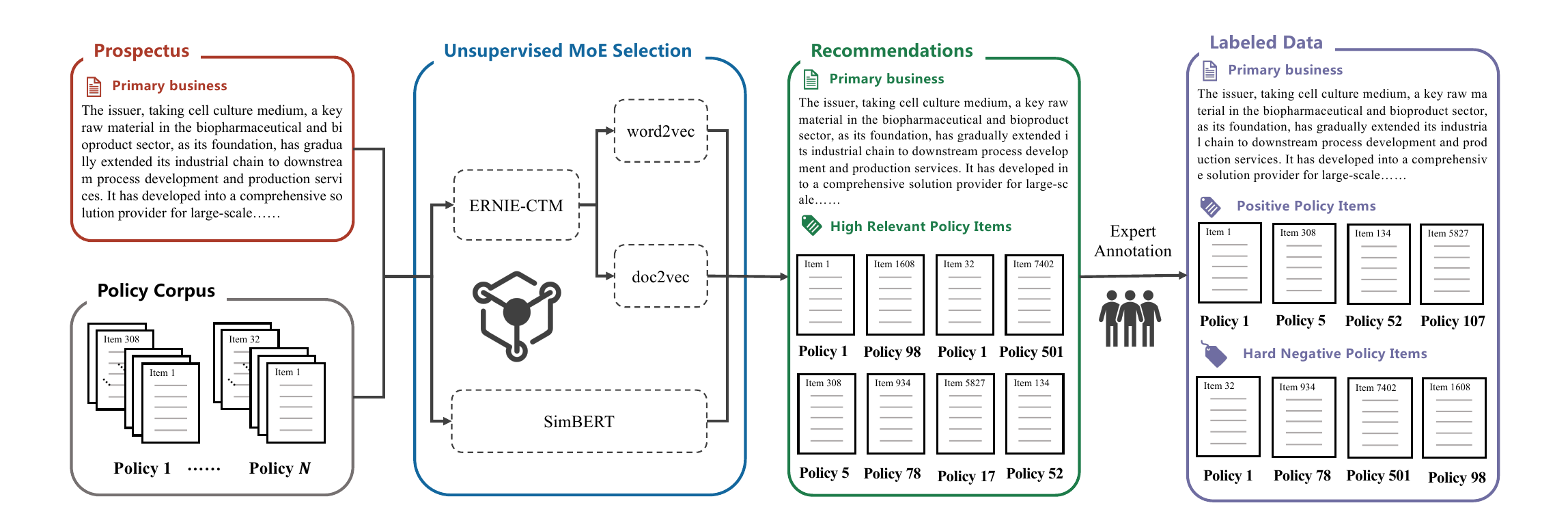}
    \caption{Overview of the annotation process. After data processing, the collected prospectus passages and policy articles are fed to a mixture-of-experts (MoE) selection system composed of unsupervised models. The Top-20 ranked policy articles for each prospectus passage are selected as recommendation for the human annotation process.}
    \label{fig:annotation}
\end{figure*}


\section{The Policy Retrieval Dataset for Stock Market in China}
\label{sec:dataset}

In this section, we detail the process of creating the CSPRD dataset, which includes the following five stages.

\subsection{Data collection}

In June 2019, the STAR Market was officially established by SSE, where listed companies are certainly conform to some incentive policy articles of Chinese government due to the requirement of SSE. From the public information on the official website\footnote{\url{https://kcb.sse.com.cn/}} of the STAR Market, we collect the prospectuses of 767 listed companies, as well as 400+ incentive policy documents compiled by SSE (as of August, 2022). The policy documents are classified into 7 categories by SSE, six of which are the exact same as the STAR Market, and the other one category is \textit{General policies}.

\subsection{Data processing}

For prospectuses, we solely extract the textual information concerning key products and services. We conduct semantic-based keyword matching and positioning through file metadata information, and specifically extract the text content under the corresponding chapter title as contextual information based on reference keywords such as \textit{main products} and \textit{primary business}. 

As for policy documents, we use regular expression to match the policy names and split them into policy articles by article title. We manually set block words to filter out political-relevant and financial-irrelevant articles.

\subsection{Unsupervised MoE Selection}

For annotation, each prospectus passage is paired with each policy article, which is in total 7 million pairs to be scored. To reduce cost of human resources, we deploy a mixture of experts (MoE) selection system to directly score the textual similarity of each pair of prospectus passage and policy content. 

Our MoE selection system is consisted of unsupervised models trained on both the policy corpus and prospectus passages. We first adopt ERNIE-CTM \cite{zhao2020TermTree} to recognize named entities in the passages and only keeps the named entities with manually selected tags. Then the named entities are used to train a \texttt{word2vec} model and a \texttt{doc2vec} model, which are used to score the textual similarity. In addition to that, we employ a pre-trained SimBERT\footnote{\url{https://github.com/PaddlePaddle/PaddleNLP} } \cite{simbert} to directly score the similarity. At the rear of the system, the final score for each text pair is the weighted sum of the scores given by \texttt{word2vec} (10\%), \texttt{doc2vec} (20\%) and SimBERT (70\%). As recommendation, we choose the 20 top-ranking policy articles for each prospectus passage.

\subsection{Expert annotation}

The CSPRD is annotated by 5 experienced SSE experts, who have systematically studied the manual \textit{QAs on the Review of Stock Issuance and Listing on the SSE STAR Market}\footnote{\url{http://www.sse.com.cn/lawandrules/sserules/tib/review/c/4729640.shtml}}. During annotation, each expert is required to focus on the primary products, main business and specific core technologies in the prospectus passage and judge whether they are compliant with the specific industry and target applications in the recommended policy articles. After cautious reading and thorough judgement, the experts should choose one of the ternary labels (\texttt{[Yes]}, \texttt{[No]}, \texttt{[Uncertain]}). Each policy pair is independently labeled by one expert and policy pairs labeled with \texttt{[Uncertain]} will be re-collected and re-labeled through group discussion of all experts. After expert annotation, the policy articles labeled as \texttt{[Yes]} are positive articles, while those labeled as \texttt{[No]} are referred as hard negative ones for contrastive learning.

\subsection{Dataset Release}

CSPRD contains a Chinese policy corpus of 10,002 articles and 709 prospectus examples from 545 companies listed on the STAR Market in China. CSPRD is bilingual in Chinese and English: the origin language of CSPRD is simplified Chinese, and the English version is translated by ChatGPT\footnote{\url{https://openai.com/blog/chatgpt}} with direct prompting. The English version is for research purpose only, the translation quality has no assurance from authors. We select 80\% data of each category as the train set, while the remaining 144 examples as the dev set.


\section{Experiments}
\label{sec:models}

 In this section, we report the performance of lexical models, embedding models and fine-tuned PLMs on CSPRD dataset as baselines for future works. We evaluate the retrieval performance with four commonly used metrics for information retrieval: mean reciprocal rank (MRR@10), normalized discounted cumulative gain (NDCG@10), recall (R@10) and precision (P@10).

\subsection{Models}

Given an example pair $(P, A)$ and a policy corpus $\mathcal{C}$ , where $P$ is the prospectus passage, $A$ is the policy article. We define the relevance score $r(P, A)$ for each method below.

\textbf{Lexical Methods}: For lexical methods, the relevance score is defined as the sum over the passage terms:
\begin{equation}
    r(P, A) = \sum\limits_{t \in P} w(t, A)
\end{equation}

We calculate the weight with TF-IDF and BM25 \cite{robertson1995okapi} approaches respectively.

\textbf{Embedding Methods}: For embedding methods, the relevance score is defined as the cosine similarity of the embeddings:
\begin{equation}
    r(P, A) = CosineSimilarity (E(P), E(A))
\end{equation}
where $E(\cdot)$ is the embedding model.

We use the texts in CSPRD dataset to fit a \texttt{word2vec} (W2V-CSPRD) and a \texttt{doc2vec} (D2V-CSPRD) models and test their performance on the dev set. In addition to that, we also benchmark an open-sourced W2V embedding model (W2V-Finance\footnote{\url{https://github.com/Embedding/Chinese-Word-Vectors}}) trained on finance texts.

\textbf{Fine-tuning Methods}: The relevance score of embedding methods is defined as the softmax score of the inner product of the encoding matrices:

\begin{equation}
    r(P, A) = Softmax_{A \in \mathcal{C}} (E(P) \cdot E(A)^T)
\end{equation}
where $E(\cdot)$ is the encoding function of the fine-tuned models.

We implement a bi-encoder paradigm with different PLMs in size of BERT$_{base}$ \cite{devlin2018bert} as encoder and fine-tune them on the train set of CSPRD. Our models are trained with in-batch negative contrastive learning proposed in DPR \cite{karpukhin2020dense}. 

Since the open-sourced RetroMAE \cite{liu2022retromae} is pre-trained on English corpus, we pre-train a RetroMAE model from scratch on $\sim$60GB publicly collected Chinese corpus and then fine-tune it on the train set of CSPRD dataset. We adopt the pre-trained Chinese BERT \cite{cui-etal-2020-revisiting} as encoder, which was pre-trained with whole word masking (WWM) \cite{cui-etal-2020-revisiting} on extra $\sim$5.4B words of Chinese corpus. During our pre-training, we keep the consistency of WWM strategy.  In pre-training task, the model is pre-trained for 5 epochs with learning rate of $1e^{-4}$ and weight decay of $0.01$. During fine-tuning, each model is fine-tuned for 10 epochs with learning rate of $2e^{-5}$.

\begin{table}[t]
\centering
\small
\caption{Retrieval benchmark of several approaches on CSPRD dev set. We pre-trained RetroMAE \cite{liu2022retromae} from scratch on $\sim$60GB Chinese corpus with Chinese BERT \cite{cui-etal-2020-revisiting} encoder. The models with $^\dagger$ are fine-tuned with DPR \cite{karpukhin2020dense} framework.}
\vspace{10pt}
\begin{tabular}{lrrrrr}
\hline
\textbf{Model}                          & \textbf{MRR@10} & \textbf{NDCG@10} & \textbf{R@10} & \textbf{P@10}\\
\hline
TF-IDF                                  &  3.2    & 1.6     & 2.2  & 1.6  \\
BM25 \cite{robertson1995okapi}          &  22.3   & 13.1    & 14.3 & 13.1   \\
\hline 
W2V-CSPRD                               &  9.9   & 11.2    & 5.3 & 18.6   \\
D2V-CSPRD                               &  10.3   & 5.2    & 6.0 & 5.2   \\
W2V-Finance                             &  19.8   & 9.9    & 10.4 & 9.9   \\
\hline
BERT$^\dagger$ \cite{devlin2018bert, cui-etal-2020-revisiting}    &  53.6   & 26.9    & 35.8 & 79.2   \\
MacBERT$^\dagger$ \cite{cui-etal-2020-revisiting}    &  50.4   & 25.4    & 34.7 & 79.2   \\
Mengzi$^\dagger$ \cite{zhang2021mengzi}    &  52.1   & 26.2    & 35.6 & \textbf{82.6}   \\
RetroMAE$^\dagger$ \cite{liu2022retromae}  &  54.8   & 27.1    & 35.8 & 79.9   \\
CoSENT$^\dagger$ \cite{text2vec}           &  \textbf{56.1}   & \textbf{28.5}    & \textbf{37.5} & 80.6   \\


\hline
\end{tabular}

\label{tab:results}
\vspace{-10pt}
\end{table}

\subsection{Results and Analysis}

Our experiment results are shown in Table \ref{tab:results}. As de facto standard methods, lexical methods TF-IDF and BM25 show poor performance on our CSPRD dataset, suggesting the challenge of our proposed policy retrieval task.

We discover that there is a positive correlation between policy relevance and textual similarity of policy articles and prospectus passages. However, models that exhibit good performance in textual similarity, without further fine-tuning, still fail to achieve satisfactory results than fine-tuned models.

Traditional methods perform rather poorly, indicating that attempting to address this task purely from the statistics of term frequency is quite challenging. Large language models (LLMs) are decoder-only models, while such task requires strong encoding capability, therefore, LLMs are not suitable for this task. For this particular task, fine-tuning smaller models is still necessary and efficient.

\section{Conclusion}

In this paper, we introduce the Chinese Stock Policy Retrieval Dataset (CSPRD), a compilation of over 700 prospectus passages accompanied by pertinent policy articles meticulously annotated by experts from the Shanghai Stock Exchange. We assessed numerous information retrieval baselines, demonstrating the utility and promise of CSPRD dataset. Our work bridges a notable gap in the realm of financial datasets for NLP and paves way for future study on policy retrieval task.

\vfill\pagebreak

\bibliographystyle{IEEEbib}
\bibliography{refs}

\appendix

\section{Data Source}
\label{sec:data_source}

Table \ref{tab:policy_source} shows the file source of CSPRD. After text extraction and filtering, 10,002 articles from 390 policies are left in CSPRD policy corpus, 709 prospectus passages of 545 listed companies are left in CSPRD train and dev sets.

\begin{table*}[htbp]
\centering
\begin{tabular}{lll}
\hline
\textbf{Source File}                          & \textbf{URL}                        &  \\
\hline
General Policies                                    & \url{http://kcb.sse.com.cn/kczl/ty/}      &  \\
Policies of New Generation of Science and Technology   & \url{http://kcb.sse.com.cn/kczl/xydxxjs/} &  \\
Policies of High-end Equipment                          & \url{http://kcb.sse.com.cn/kczl/gdzb/}    &  \\
Policies of New Materials                              & \url{http://kcb.sse.com.cn/kczl/xcl/}     &  \\
Policies of New Energy                                 & \url{http://kcb.sse.com.cn/kczl/xny/}     &  \\
Policies of Environment Protection                      & \url{http://kcb.sse.com.cn/kczl/jnhb/}    &  \\
Policies of Biomedicine                                 & \url{http://kcb.sse.com.cn/kczl/swyy/}    &  \\
Prospectus of listed Companies                          & \url{http://kcb.sse.com.cn/disclosure/}  & \\
\hline
\end{tabular}
\caption{Source files from SSE STAR Market website}
\label{tab:policy_source}
\end{table*}

\begin{figure*}[htbp]
    \centering
    \includegraphics[scale=0.4]{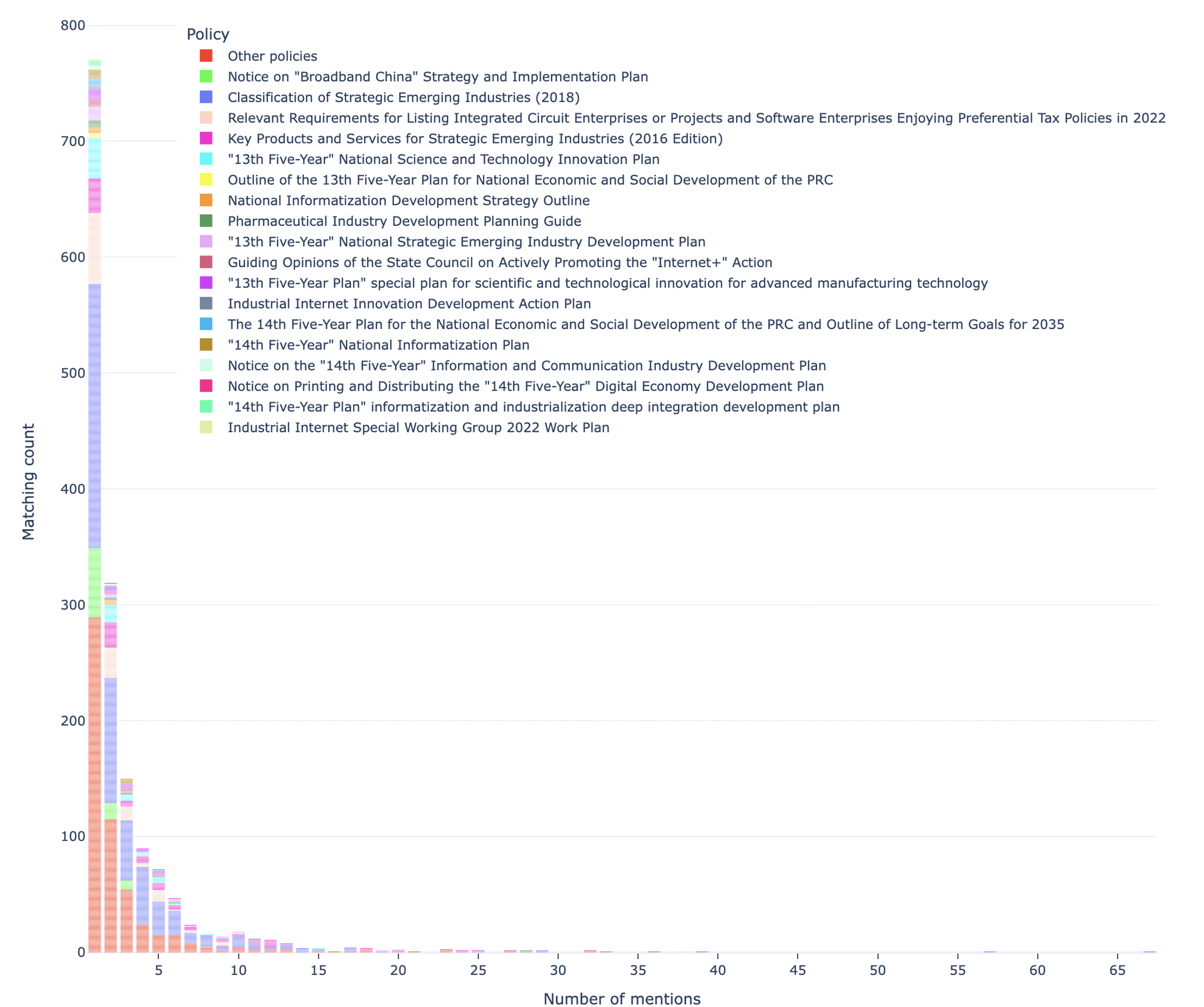}
    \caption{Distribution of the number of matched prospectus passages per policy article. }
    \label{fig:dataset5}
\end{figure*}

\section{Dataset Statistics}
\label{app:sta}

\begin{figure*}[htbp]
    \centering
    \includegraphics[scale=0.25]{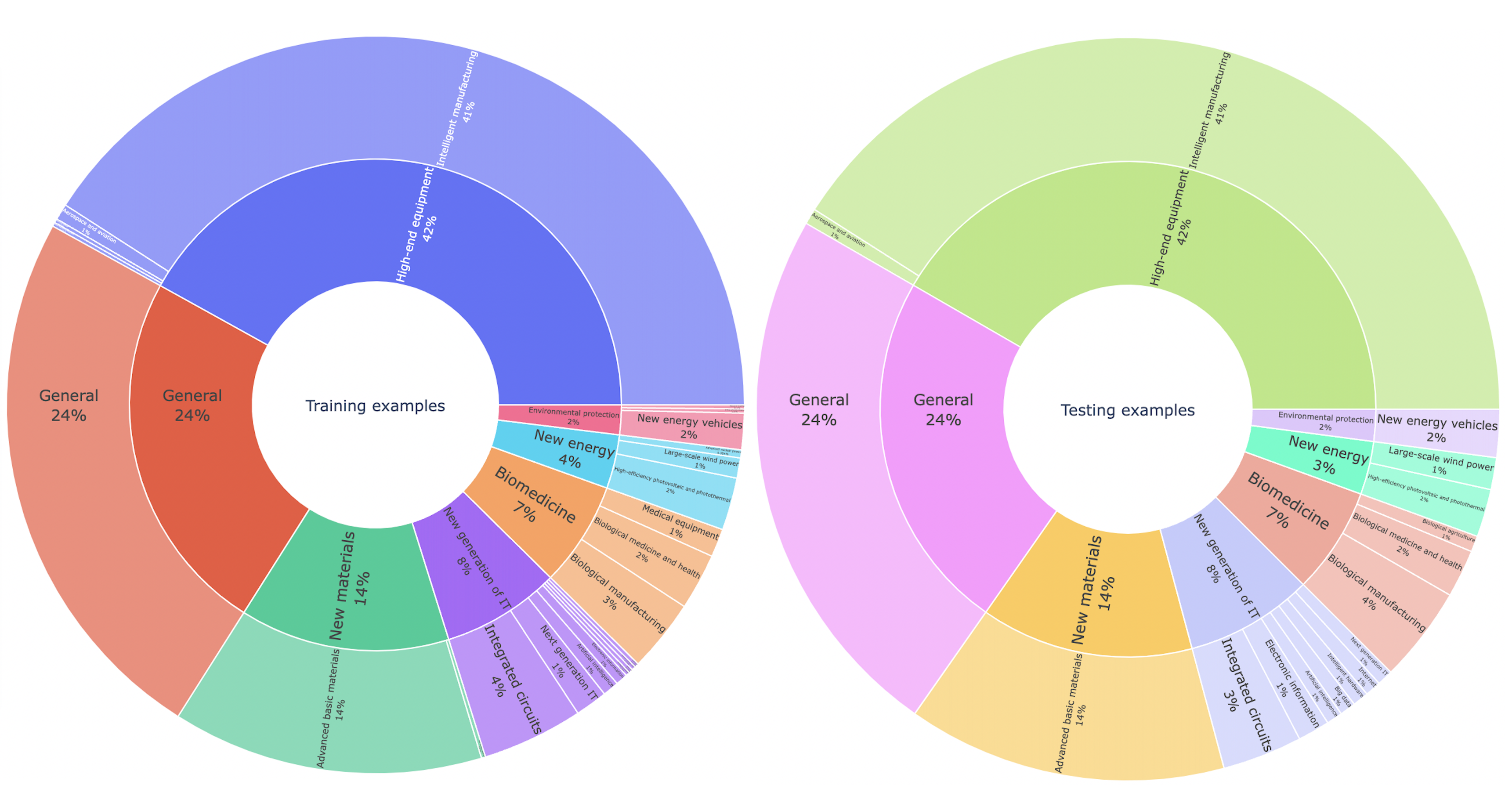}
    \caption{Statistics of the samples in the train and dev sets of CSPRD. CSPRD samples are labeled into seven categories, six of which are the exact same as the STAR Market categories, and the other one category is \textit{General}.}
    \label{fig:dataset}
\end{figure*}

\begin{figure*}[htbp]
    \centering
    \includegraphics[scale=0.3]{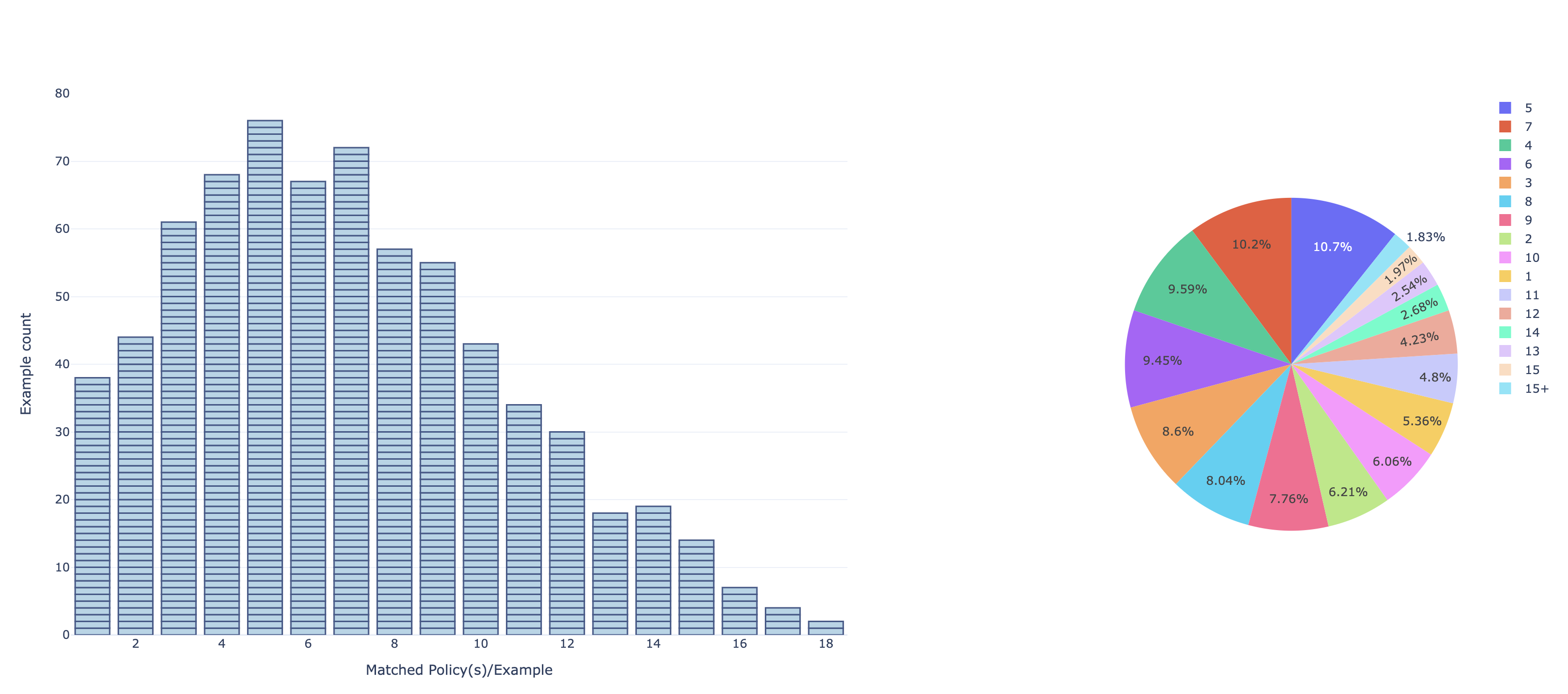}
    \caption{Statistics of the number of relevant policy articles per prospectus passage}
    \label{fig:dataset2}
\end{figure*}

\begin{figure*}[htbp]
    \centering
    \includegraphics[scale=0.3]{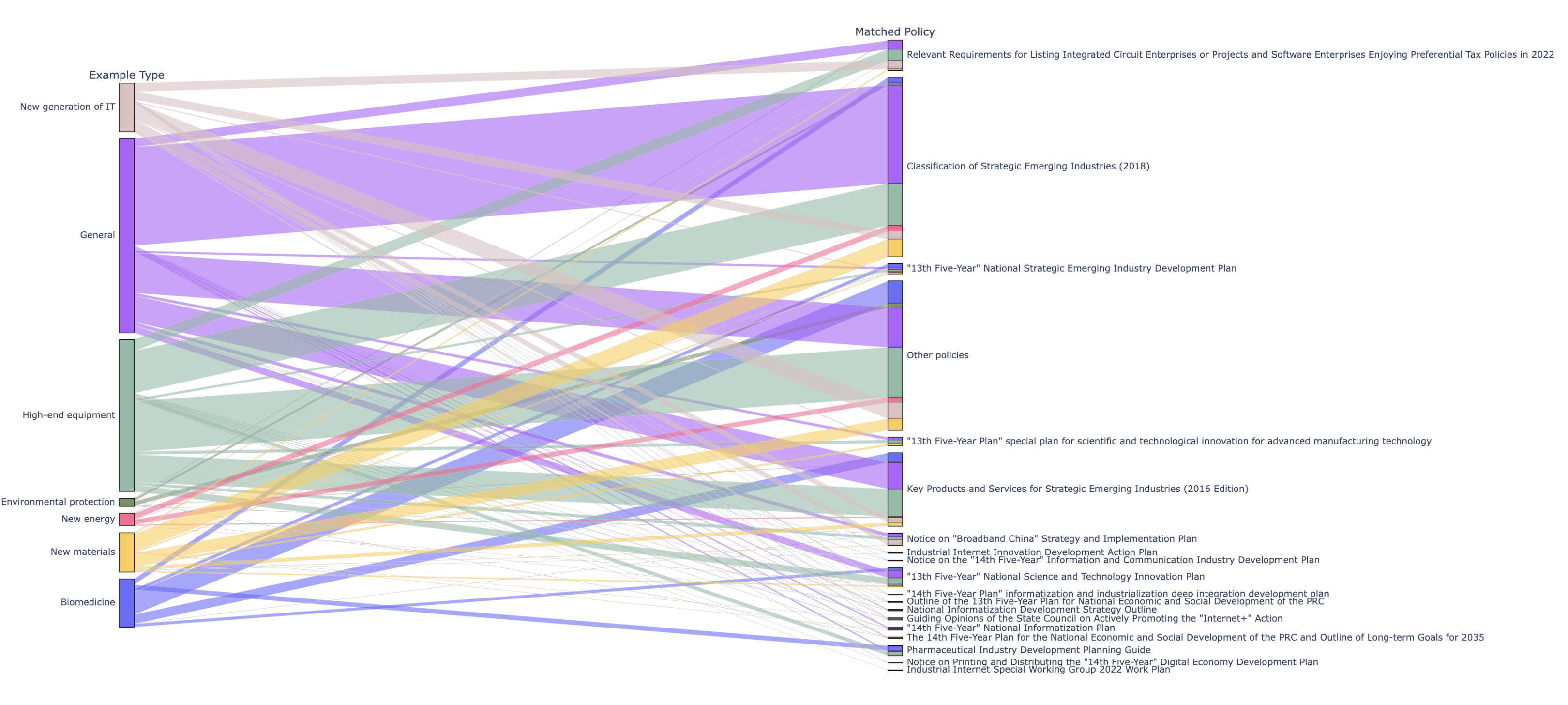}
    \caption{Matching distribution between prospectus passage and Top20 policy documents}
    \label{fig:dataset3}
\end{figure*}

\begin{figure*}[htbp]
    \centering
    \includegraphics[scale=0.3]{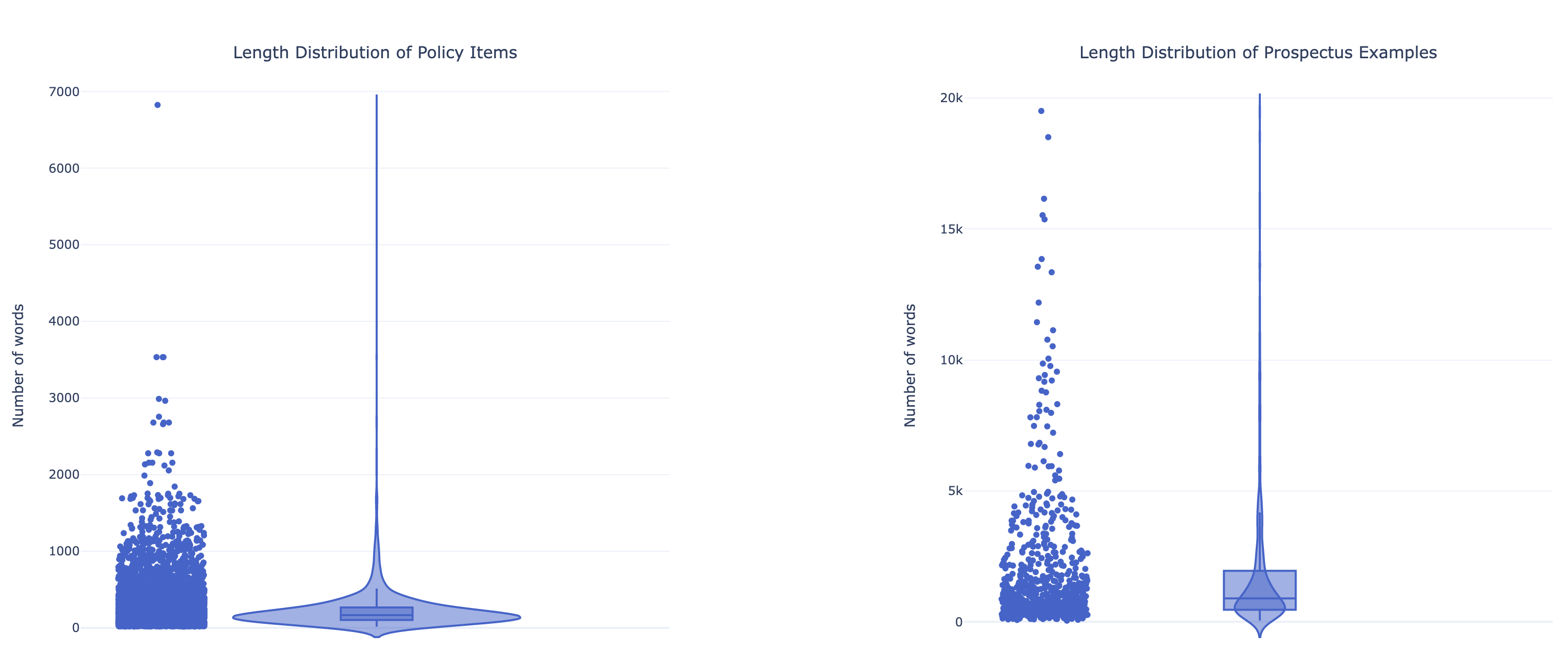}
    \caption{Length distribution of prospectus passages and policy articles}
    \label{fig:dataset4}
\end{figure*}

We visualize our dataset statistics in Figure \ref{fig:dataset5}, \ref{fig:dataset}, \ref{fig:dataset2}, \ref{fig:dataset3} and \ref{fig:dataset4}.

\section{Dataset Translation}
\label{app:trans}

The English version of CSPRD is translated by ChatGPT (\texttt{gpt-3.5-turbo-16k-0613}) via OpenAI API. The English version is for research purpose only, the translation quality has no assurance from authors. The prompt for translation is: 

\begin{itemize}
 \item \texttt{Translate the following Chinese text to English:\textbackslash n'\{text\}'}
\end{itemize}
The temperature is set to 0 for less randomness.

\section{Experiment Setting}

The base models from Huggingface Hub are respectively:
\begin{itemize}
    \item BERT \cite{devlin2018bert, cui-etal-2020-revisiting}: \texttt{hfl/chinese-bert-wwm-ext}
    \item MacBERT \cite{cui-etal-2020-revisiting}: \texttt{hfl/chinese-macbert-base}
    \item Mengzi \cite{zhang2021mengzi}: \texttt{langboat/mengzi-bert-base-fin}
    \item RetroMAE \cite{liu2022retromae}: \texttt{hfl/chinese-bert-wwm-ext}
    \item CoSENT \cite{text2vec}: \texttt{shibing624/text2vec-base-chinese}
\end{itemize}



\end{document}